\documentclass[runningheads]{llncs}

\usepackage{eccv}

\usepackage{eccvabbrv}
\usepackage{multirow}
\usepackage{colortbl}
\usepackage{graphicx}
\usepackage{booktabs}

\usepackage[accsupp]{axessibility}  

\usepackage{hyperref}

\usepackage{orcidlink}
\newcommand{\methodName}{MarvelOVD\xspace}

\begin{document}

\title{MarvelOVD: Marrying Object Recognition and Vision-Language Models for Robust Open-Vocabulary Object Detection} 

\titlerunning{Marrying Object Recognition and Vision-Language Models}

\author{Kuo Wang\inst{1}\orcidlink{0000-0003-2522-4134} \and
Lechao Cheng\inst{2}$^\star$ \and
Weikai Chen\inst{3} \and
Pingping Zhang\inst{4} \and
Liang Lin\inst{1,5} \and
Fan Zhou\inst{6} \and
Guanbin Li\inst{1,5}\thanks{Corresponding authors.}
}
\authorrunning{K.~Wang et al.}

\institute{School of Computer Science and Engineering, Sun Yat-sen University\and
Hefei University of Technology \and
Tencent America \and
Dalian University of Technology \and
Peng Cheng Laboratory \and
School of Computer Science and Engineering, National Engineering Research Center of Digital Life, Sun Yat-sen University
\\
\email{wangk229@mail2.sysu.edu.cn, chenglc@hfut.edu.cn, chenwk891@gmail.com, zhpp@dlut.edu.cn, linliang@ieee.org, \{isszf, liguanbin\}@mail.sysu.edu.cn}
}
\maketitle


\begin{abstract}
Learning from pseudo-labels that generated with VLMs~(Vision Language Models) has been shown as a promising solution to assist open vocabulary detection (OVD) in recent studies. However, due to the domain gap between VLM and vision-detection tasks, pseudo-labels produced by the VLMs are prone to be noisy, while the training design of the detector further amplifies the bias. In this work, we investigate the root cause of VLMs' biased prediction under the OVD context. Our observations lead to a simple yet effective paradigm, coded \methodName, that generates significantly better training targets and optimizes the learning procedure in an online manner by marrying the capability of the detector with the vision-language model. Our key insight is that the detector itself can act as a strong auxiliary guidance to accommodate VLM's inability of understanding both the ``background'' and the context of a proposal within the image. Based on it, we greatly purify the noisy pseudo-labels via Online Mining and propose Adaptive Reweighting to effectively suppress the biased training boxes that are not well aligned with the target object. In addition, we also identify a neglected ``base-novel-conflict'' problem and introduce stratified label assignments to prevent it. Extensive experiments on COCO and LVIS datasets demonstrate that our method outperforms the other state-of-the-arts by significant margins. Codes are available at \url{https://github.com/wkfdb/MarvelOVD}.

\end{abstract}

\section{Introduction}
\label{sec:intro}

Open-vocabulary object detection (OVD)~\cite{zareian2021open} is receiving increasing attention due to its capability of detecting novel objects at test time. In a typical OVD setting, only a fraction of the target categories is annotated (referred as the base categories) while the goal of OVD is to recognize a set of novel classes at inference time. The objects of novel categories can appear in the training images but do not receive any annotations.
To enhance the generalizability of the OVD detector, recent works have proposed to incorporate the vision-language models (VLMs)~\cite{radford2021learning, jia2021scaling,zhai2022lit}, which have been verified with excellent zero-shot recognization capacity, to improve the existing OVD pipeline. 

\begin{figure}[t]
  \centering
    \includegraphics[width=\linewidth]{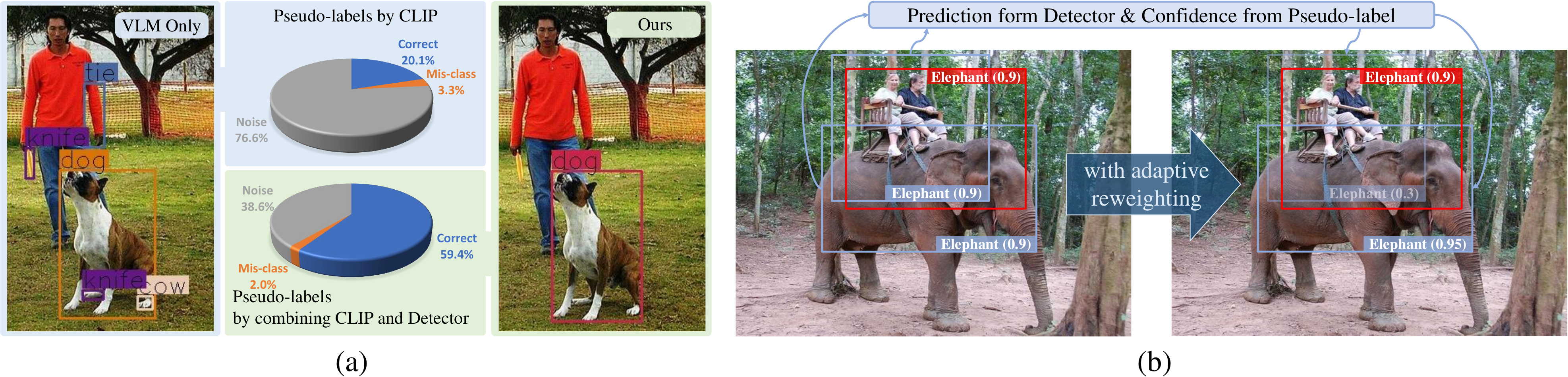}
  \caption{Improvements achieved by incorporating the detector for pseudo-label generation and the following learning phase. \textbf{(a)} The distribution of pseudo-labels generated by CLIP and our method. ``Mis-class'' means boxes labeled as wrong categories and ``noise'' indicates boxes that should not be considered as pseudo-labels. The VLM~(CLIP) has low ``mis-class'' rate but fails to distinguish noisy boxes. Our method discriminates the noises by combining the characteristics of the detector, and hence significantly improves the quality of the pseudo labels. \textbf{(b)} The red box indicates pseudo-label and the blue boxes represent the matched training boxes. Adaptive proposal reweighting computes independent weights according to the prediction of detector and the confidence from pseudo-label, leading the training to focus on more reliable instances (e.g. the lower right training box).}
  \label{fig1}

\end{figure}

A common practice for the OVD task with known novel concepts is to generate pseudo-labels using the VLMs~(e.g. CLIP~\cite{radford2021learning}) in an offline manner~\cite{zhao2022exploiting}. However, because of the domain shift between the contrastive language-image pretraining and object detection tasks, VLMs trained with image-level data inevitably introduce noisy annotations when applied to the cropped partial images. We demonstrate an in-depth analysis of the pseudo-labels generated by the CLIP model in Figure~\ref{fig1}(a). We denote the valid novel object proposal mistakenly classified as other categories as \textit{``Mis-class''} and boxes that should not be considered to contain a novel object as \textit{``Noise''}. In fact, the mis-classification rate of the VLM-based method is rather low (only $3.3\%$). The main source of error stems from its incapability of distinguishing ``noisy'' boxes (error rate $76.6\%$), e.g., the dog leg in Figure~\ref{fig1}(a), that should not be considered as a valid object of interest.

The key reasons for the difficulty of the VLMs to recognize noisy proposals are two-fold. 
1) The lack of contextual information to understand the locally cropped images. 
Instead of trained with image patches, the CLIP model is fed with complete images with paired texts.
Therefore, it is not able to leverage the image context outside the input proposal, which may be crucial to interpret the semantics of the candidate box. For instance, in Figure~\ref{fig1}(a), the man's arm is falsely classified by the CLIP model as ``tie'', as it fails to recognize that the seemingly ``tie'' object is in fact connected to a human body. 
2) The unawareness of the ``background'' elements. The CLIP model generates the category prediction by computing the similarity between the query image feature and the text embedding of the candidate categories. Since ``background'' is relatively defined according to interested foreground categories, there is no pre-defined text embedding to represent the concept of ``background'' during the inference. However, the CLIP model still has to provide a prediction even when the input content is not related to any of the target categories.
In Figure~\ref{fig1}(a), the dog leg falls in this case -- it is identified as ``cow'' only because it appears more like cow than any other category, leading to noisy boxes.

Unlike the CLIP model, the RoI align technique in detectors naturally provides rich contextual information for local regions. Moreover, the detector is aware of the  concept of ``background'' during inference. As a result, the noisy boxes that confuse VLMs can be clarified by the detector as ``background'' with high confidence. Inspired by this key observation, we propose \methodName, a dedicated framework for open-vocabulary detection that can yield high-quality pseudo-labels and marvelous performance by combining the merits of the object detector and the vision-language models. In particular, we leverage the ``context and background'' awareness of the detector as strong auxiliary guidance to comprehensively improve the \textit{pseudo-label generation} pipeline and the \textit{training} procedure.

For pseudo-label generation, the predicted category confidence is based on a weighted sum of the outputs from the detector and the VLM, favoring the reliable classification of the VLM models while ruling out noisy boxes using the detector. To accelerate the training, we pre-generate the VLM predictions on all the candidate boxes and dynamically mine credible pseudo-labels under the guidance of the detector at each training iteration. The complementary capabilities of the detector and the VLM significantly improve the accuracy of pseudo-labels, even at the early training stage. Moreover, as the detector improves during training, the quality of the generated pseudo-labels increases as well, which eventually boosts the final performance as shown in Figure~\ref{fig1}(a).

Conventional training design of object detectors~\cite{he2017mask} equally treats each proposal that matches with one training target. Such a design is not suitable for learning from pseudo-labels. Specifically, the generated pseudo-box may deviate a lot from the bounding box of the real novel object. Therefore, as shown in Figure~\ref{fig1}(b), the overlaps between the training boxes and the actual novel object usually present a large variance. 
This means that these training boxes should not make equal contributions to the final loss, even if they match the same pseudo-label. To this end, instead of weighting the pseudo-labels~\cite{zhao2022exploiting}, we adaptively compute individual weights for each training box that is matched with a pseudo-label. As shown in Figure~\ref{fig1}(b), training boxes with inaccurate positions will receive smaller weights and vice versa. Note that the training boxes are produced with a stratified label assignment strategy, which eliminates the conflicts between the pseudo-labels and base annotations and thus prevents the negative influence of noisy pseudo-label on the performance of base category detection. We conduct extensive experiments on COCO and LVIS datasets. \methodName consistently outperforms the state-of-the-art methods by a great margin. In summary, our contributions are as follows.
\begin{itemize}
     \item We identify the fundamental causes of the VLM's biased prediction and provide in-depth analysis.
     \item We propose \methodName, a novel OVD framework that generates high-quality pseudo-labels by leveraging the detector as auxiliary guidance to mitigate the domain gap of VLMs on image patch prediction.
     \item A novel proposal re-weighting mechanism and the stratified label assignment method to improve the OVD training performance.
    \item We set the new state of arts in the MS-COCO and LVIS benchmarks.
 \end{itemize}

\section{Related Work}
\paragraph{Vision-Language Pre-training} 
Vision-Language pre-training, aiming to align visual and textual representations, employs contrastive learning on large-scale image-caption pairs~\cite{frome2013devise,jayaraman2014zero,kim2021vilt}. Significant research has enhanced various downstream tasks using Vision-Language models~\cite{li2022blip,li2021align,lu2019vilbert}. Notably, models like CLIP~\cite{radford2021learning} and ALIGN~\cite{jia2021scaling} have leveraged billion-scale image-text pairs for vision-language representation learning, achieving remarkable success in zero-shot image classification and image-text retrieval. This success has inspired the application of Vision-Language Models (VLMs) in enhancing the range and accuracy of dense recognition tasks, such as object detection~\cite{wang2023object,wu2023cora,gu2021open,shi2022proposalclip} and semantic segmentation~\cite{yun2023ifseg,zhou2022extract,rao2022denseclip}. Nonetheless, VLMs, typically pre-trained by considering the entire image, face domain gaps in dense prediction tasks. Our research investigates the application of VLMs for object detection in localized regions, aiming to broaden the detector's cognitive scope without relying on manual annotations.
\paragraph{Open Vocabulary Object Detection} Open Vocabulary Object Detection (OVD) aims to extend the detector's recognition capability to classes not present in the training data using auxiliary data or models. The concept was initially introduced by OVR-CNN~\cite{zareian2021open}, which empowered the detector to recognize diverse object concepts by leveraging vision and text encoders pre-trained on image-text pairs. Subsequently, several OVD methods have been developed. Recent research in OVD has explored various forms of auxiliary data, including transfer learning with image-text pairs~\cite{lin2022learning,bravo2022localized,chen2022open}, knowledge distillation from pre-trained Vision-Language Models~\cite{wang2023object,wu2023aligning,du2022learning,bangalath2022bridging,gu2021open}, pseudo-label generation from image classification data~\cite{zhou2022detecting,bangalath2022bridging}, and pretraining with grounding data~\cite{li2022grounded,zhang2022glipv2}. Except for CNN-based detectors, transformer-based open vocabulary detectors~\cite{li2023distilling, zang2022open, wu2023cora} have also been widely exploited. In addition to the standard OVD scenario, which assumes that novel categories are unknown during training, some existing works~\cite{zhao2022exploiting,feng2022promptdet} have also explored open vocabulary detection with prior knowledge about potential novel concepts. Typically, in such cases, pseudo-labels for novel categories are generated using Vision-Language Models (VLMs) before training. However, the domain gaps between vision-language pre-training and object detection introduce noisy pseudo-labels, significantly constraining the performance of existing methods. To address this challenge, we identify a critical limitation in applying VLMs to localized regions and propose a solution that involves integrating the detector's capabilities to effectively mitigate this noise, leading to substantially improved performance.

\section{Preliminaries}

Open-vocabulary object detection aims to train detection models by leveraging a dataset $\mathcal{D}=\{\boldsymbol{x}_{i}, y_i\}_{i=1}^{n}$ and auxiliary weakly supervised data~(e.g. image-text pairs, VLMs, etc.), where $\boldsymbol{x}_{i}$ represents the image and $y_i$ includes the location and category of the objects contained in the image. Different from conventional detection tasks, the annotation of images only covers the base categories $\mathcal{C}^{B}$, while the OVD task requires the detector to additionally detect novel categories $\mathcal{C}^{N}$ at the test time. Note that $\mathcal{C}^{B}\cap\mathcal{C}^{N}=\varnothing$, and the label space $\mathcal{C}^{N}$ is already known during the training.

To achieve the detection in an open-vocabulary label space, the classification head in the detector is designed to compare the similarity between the region-based visual embeddings and the text embeddings~\cite{zareian2021open}. In particular, the region-embeddings $\boldsymbol{R}=\{\boldsymbol{r}_i\}_{i=1}^{N_r}$ are obtained through RoI Align and the following feature extractor, where $N_r$ is the number of regional boxes in the images. The text embeddings are composed of $\boldsymbol{C}=\{\boldsymbol{c}_{bg}\}\cup\{\boldsymbol{c}_i\}_{i=1}^{N_c}$, where $\boldsymbol{c}_i$ is obtained by feeding the category name with template prompts~(e.g., {``\tt{a photo of \{category name\} in the scene}''}) to the pre-trained text encoder and $N_c$ is number of categories. The $\boldsymbol{c}_{bg}$ is initialized as a learnable embedding. Based on the regional and text embeddings, the probability of the region $\boldsymbol{r}_i$ being classified as category $\boldsymbol{c}_j$ is defined as  
\begin{equation}
p_{i,j} = \frac{exp(\boldsymbol{r}_i\cdot\boldsymbol{c}_j)}{exp(\boldsymbol{r}_i\cdot\boldsymbol{c}_{bg})+\sum_{k=1}^{N_c}exp(\boldsymbol{r}_i\cdot\boldsymbol{c}_k)}.
\label{eq1}
\end{equation}
Comparing the similarity between the region and text embedding enables the detector to recognize objects in an unlimited label space. 

\section{MarvelOVD}
\begin{figure*}[t]
  \centering
  \includegraphics[width=\linewidth]{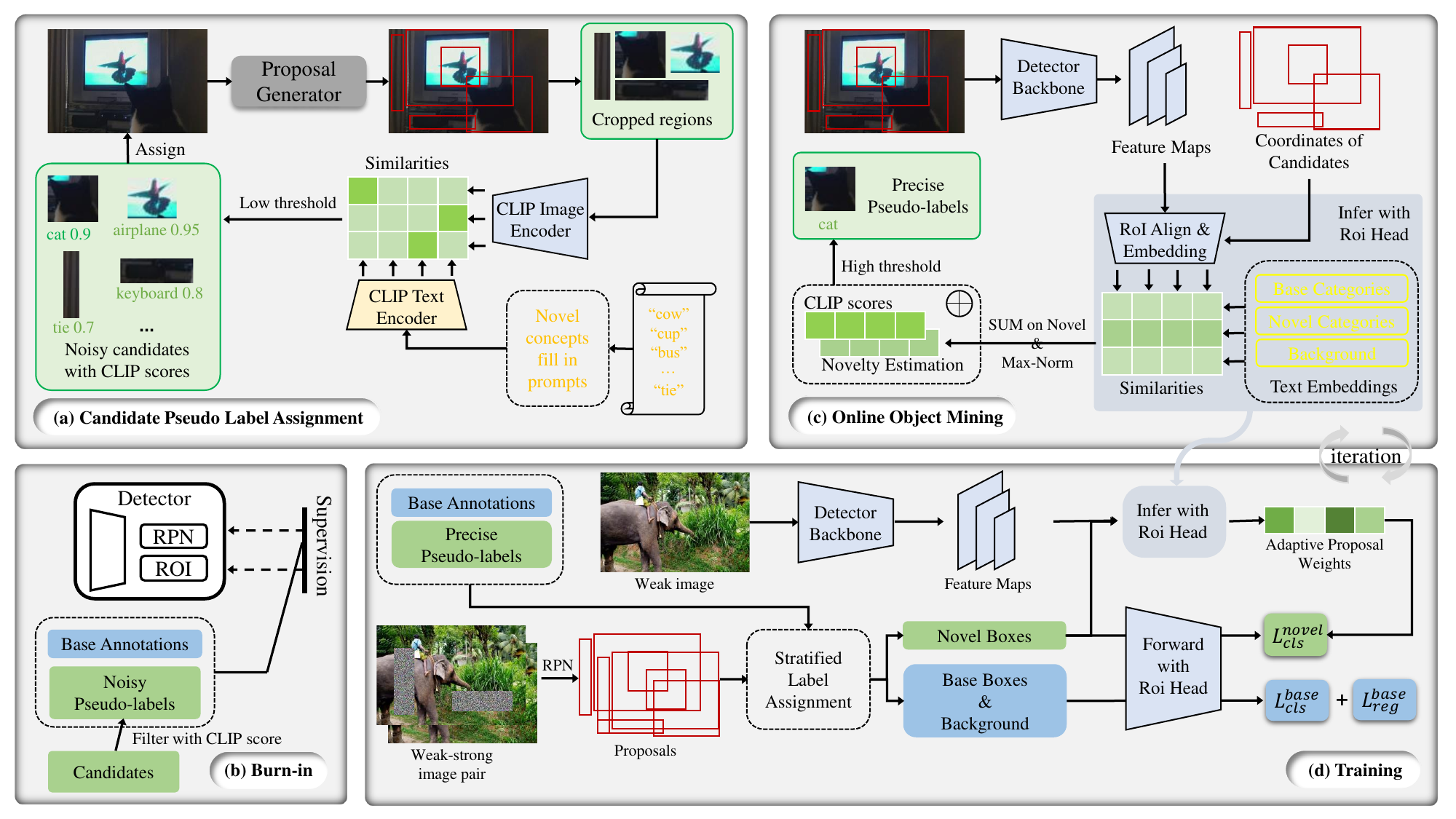}
  \caption{The framework of our method, which improves the quality of pseudo-labels while optimizing the following learning process by dynamically incorporating the detector during the training. We first assign candidate boxes to the images with CLIP and a proposal generator. Then we select noisy pseudo-labels according to the CLIP scores to burn-in the detector. After burn-in, the detector initially obtains the capacity to recognize novel concepts. Based on it, we dynamically estimate the novelty of each candidate box and combine the corresponding CLIP prediction to select precise pseudo-labels. We adopt stratified label assignment to generate training boxes, while the loss weights for the novel training boxes are independently computed based on the detector's prediction.}
  \label{fig2}
\end{figure*}

To facilitate the learning of semantics associated with open categories without manual annotation, existing approaches often employ pre-trained Vision-Language Models (VLMs) to discover potential novel objects~\cite{zhou2022detecting,zhao2022exploiting,gao2022open,feng2022promptdet} and generate pseudo-labels for subsequent training. The typical procedure involves training a proposal generator using base annotations to identify localized regions that may contain novel objects, followed by the generation of pseudo-labels based on VLM inference results within these cropped regions. However, as VLMs are pre-trained on entire images, their application to localized regions inevitably introduces noisy pseudo-labels, leading to disruptions in the learning process for novel categories.
To enhance the learning of novel concepts, we present MarvelOVD, which dynamically integrates the detector's capabilities into the pseudo-label generation process while optimizing the subsequent learning stages. Figure~\ref{fig2} provides an overview of our framework, with detailed explanations in the following sections.

\subsection{Candidate Pseudo-Label Assignment}

Our proposed MarvelOVD dynamically generates better pseudo-labels at each training iteration under the guidance of both the detector and pre-trained VLMs. Since predicting the cropped regions with VLMs during training requires unaffordable time overhead, we alternatively assign candidate pseudo-labels to each image before the training and then select precise ones to train the detector at each iteration.

Following the existing method~\cite{zhao2022exploiting}, we first train a class-agnostic proposal generator with the base annotation to produce regional boxes for each image. Image patches are then cropped according to the regions and fed into the CLIP image-encoder to obtain the regional-visual-embeddings. At the same time, we utilize the corresponding CLIP text-encoder and template text prompts to encode each novel category. After that, a similarity matrix is computed via dot product to describe the similarity between each visual-embedding and text-embedding. Finally, softmax is applied to obtain the distribution over novel categories for each region. Based on it, the conventional method~\cite{zhao2022exploiting, gao2022open} post-processes the boxes and selects high-confidence pseudo-labels to train the detector. The problem is that the domain gap easily leads CLIP to produce high confidence predictions on noisy regions, which greatly limits the performance of the existing methods. In contrast, we record the CLIP predictions and assign candidate boxes to the image using a low threshold~(e.g. 0.5), and then dynamically select precise pseudo-labels from the candidates under the guidance of the detector, which is introduced in detail in the following section.

\subsection{Online Pseudo-label Mining}

The candidate pseudo-labels can be categorized into two groups: ground-truth boxes (those tightly enclosing actual novel objects) and noisy boxes (those that should not be designated as novel objects). The primary objective is to eliminate noisy candidates while retaining ground-truth boxes as reliable pseudo-labels for training. The key distinction between the detector and CLIP in their inference of localized regions lies in the contextual information and the ``background'' concept. The RoI Align mechanism employed by the detector excels at extracting contextual features from the boxes, a capability that is absent when cropping images based on coordinates alone. Additionally, the detection task incorporates a specific task-specific category known as ``background'', a concept that CLIP remains unaware of when inferring localized regions. 
Failing to recognize the ``background'' objects and the lack of contextual features are the root causes of CLIP model generating wrong predictions for noisy boxes.
Though the CLIP model struggles with the noisy boxes, 
its predictions for the ground-truth boxes are highly accurate. 
Leveraging this insight, we propose to utilize the detector's predictions for these candidates to estimate whether a box  encloses real novel objects. 
Subsequently, we combine CLIP's classification results to select high-quality pseudo-labels.

\paragraph{Burn-in.} To estimate the novelty of the candidate boxes, the detector first needs to learn ``what is novel''. To achieve this, we first utilize the top-1 CLIP score~(top score of the distribution predicted with CLIP's image-encoder and text-encoder) and a fixed threshold 0.8~(best threshold in previous work~\cite{zhao2022exploiting}) to initially select pseudo-labels to burn-in the detector for $\omega$ steps. After the burn-in phase, the model will initially gain the ability to distinguish between base objects, novel objects, and the background.

\paragraph{Online Object Mining.} Online object mining officially begins after the burn-in phase. We draw ideas from semi-supervised learning~\cite{sohn2020fixmatch} to derive weak-strong image pairs for the training, which enhances the learning for pseudo-labels. In particular, we first predict the candidate boxes with the detector on weakly augmented features. Based on it, we compute a novelty score for each candidate as follows:
\begin{equation}\label{eq2}
\begin{aligned}
z_{i} = \frac{\sum_{k\in\mathcal{C}^{N}}exp(\boldsymbol{r}_i\cdot\boldsymbol{c}_k)}{\sum_{j\in\mathcal{C}^{B}\cup\mathcal{C}^{N}\cup\{\boldsymbol{c}_{bg}\}}exp(\boldsymbol{r}_i\cdot\boldsymbol{c}_j)}
\end{aligned}
\end{equation}
where $\boldsymbol{r}$ is the vision-embedding calculated by the detector and $\boldsymbol{c}$ is the text-embedding of categories. $\mathcal{C}^B$ and  $\mathcal{C}^N$ are the sets of base and novel categories, respectively. The novelty score $z_i$ relatively estimates the novelty of candidate boxes with respect to the base category and background. However, its value varies drastically with different degrees of convergence. To tackle this problem, we further apply max-norm to the novelty scores of candidates to obtain stable estimations:
\begin{equation}\label{eq3}
s_{i}^{det} = \frac{z_i}{max\{z_1, z_2, \cdots, z_{N_r}\}} 
\end{equation}
where $N_r$ represents the number of the candidates. Benefiting from the contextual reasoning capacity of the detector and the awareness of background, the novelty estimation $s^{det}$ computed by the detector can more precisely distinguish the ground-truth/noise candidates. Combined with the accurate classification prediction generated by the CLIP model, we finally calculate the confidence score for each candidate box as follows:
\begin{equation}\label{eq4}
s_{i} = \lambda s_i^{CLIP} + (1-\lambda) s_i^{det}
\end{equation}
In the above equation, $s_i^{CLIP}$ means the top-1 CLIP score and $\lambda \in [0,1]$ is a scalar that controls the dependency of two different models. We utilize a fixed threshold $\delta$ to select high-quality pseudo-labels. The training is then derived on both weakly and strongly augmented images. 

The incorporation of the detector significantly reduces the confidence of the noisy candidates, which greatly improves the accuracy of the selected pseudo-labels, even at the initial training phase. Moreover, as the model converges with the training, the novelty estimation $s^{det}$ would be more accurate, which results in pseudo-labels of higher quality and ultimately boosts the model's detection performance on novel categories.

\subsection{Training}
In this section we describe our improvements to the conventional training design of the detector~\cite{he2017mask}. All the proposed methods in this section are applied both in the burn-in stage and the following online-object-mining stage.
\paragraph{Stratified Label Assignment.} 
The learning of novel concepts should not affect the model's performance in recognizing base objects. However, an easily overlooked phenomenon is that the mAP for base categories drops when novel pseudo-labels are applied for the training. The reason is that the novel pseudo-labels may overlap with the base annotation, resulting in 
``base-novel-conflicts'' in the IoU-based label assignment. To tackle this problem we propose stratified label assignment, which first assigns proposals with base annotations by IoU-matching, and boxes that are marked as background in the first step are secondly matched with the pseudo-labels. Experiments demonstrate that stratified label assignment helps achieve 
high accuracy of detecting novel objects without compromising the performance on estimating base categories.

\paragraph{Adaptive Proposal Reweighting.}
Since the localization quality of the pseudo-label is limited, the box center may be far away from the ground-truth object center. As a result, training boxes that matched with the mislocalized pseudo-label share extremely unbalanced overlaps with the ground-truth object. However, the conventional training design of detectors~\cite{he2017mask} equally derives training losses on those unbalanced boxes, which hampers the learning process. 
To resolve this issue, we propose adaptive proposal reweighting to assign independent loss weights to each training box that matches with pseudo-label.

The loss function to train the detector with adaptive proposal reweighting is computed as:
\begin{equation}\label{eq5}
\mathcal{L} = \frac{1}{N}(\sum_{i=1}^{n^{base}}l(b_i^{base}, \mathcal{G}^{base})+\gamma\sum_{i=1}^{n^{novel}}w_i\cdot l(b_i^{novel}, \mathcal{G}^{novel}))
\end{equation}
where $N=n^{base}+n^{novel}$ ($n^{base}$ includes background box) is the total number of training boxes, $\gamma$ is the overall weight for novel concept learning and $w_i$ represents the independent weights for each novel training box. In particular, we follow the design of Eq.~\ref{eq4} to define the individual weight $w_i$ as:
\begin{equation}\label{eq6}
w_{i} = \lambda^\prime s_i + (1-\lambda^\prime)r_i\
\end{equation}

In Eq.~\ref{eq6}, $s_i$ indicates the confidence of the corresponding pseudo-label and $r_i$ is a reliability score estimated for each matched training box. Estimating the reliability score $r_i$ is crucial and challenging. We empirically find that background score predicted on the weakly augmented images keeps close negative correlation to the overlaps with actual object and define $r_i = 1-b_i$, where $b_i$ is the background score predicted according Eq~\ref{eq1}. We also examine other indicators for comparison, more details are shown in supplementary materials. With adaptive reweighting, training boxes with higher overlap to real novel objects will be given greater weights and vice versa, thus de-biasing the learning procedure for novel concepts and further improving the performance.

\section{Experiments}

In this section, we evaluate our MarvelOVD framework against standard benchmarks, comparing it with current state-of-the-art approaches. Additionally, our ablation studies provide in-depth analyses of the primary issues leading to noises in traditional CLIP-based pseudo-label generation and detail how our framework effectively addresses these challenges.

\subsection{Datasets} 
Our primary experiments utilize the COCO-2017 dataset~\cite{lin2014microsoft} in an open vocabulary setting~\cite{zareian2021open}, dividing 48 base and 17 novel categories for evaluation. Annotations for base categories are provided, while only category names are available for novel classes. We calculate $AP_{50}^{Novel}, AP_{50}^{Base}, AP_{50}^{All}$, representing the mean Average Precision at an IoU of 0.5 for novel, base, and all categories, respectively. Additionally, the LVIS-v1 dataset~\cite{gupta2019lvis} is employed in standard Open Vocabulary Detection (OVD) settings~\cite{gu2021open}, treating 337 rare categories as novel and the rest as base. For LVIS, we report box Average Precision (AP) averaged over IoUs from 0.5 to 0.95 for rare (novel), common, frequent, and all categories, denoted as $AP_r, AP_c, AP_f$, and $AP$.

\subsection{Implementation Details} 
We utilize ViT-B/32 CLIP as the pre-trained Vision-Language Model (VLM) and its text-encoder for encoding category concepts. Consistent with existing approaches~\cite{zhao2022exploiting}, our experiments on the COCO dataset employ Mask-RCNN~\cite{he2017mask} with ResNet50-FPN~\cite{he2016deep,lin2017feature} as the base detector. For training, we initially select noise pseudo labels with CLIP scores above 0.8 and use this setup for the burn-in phase for $\omega=0.5k$ iterations. Subsequently, we set $\lambda,\lambda^\prime=0.5$ to integrate the detector with CLIP and $\delta=0.9$ for generating precise pseudo-labels. We set $\gamma=2$ as the overall weights for novel concept learning. Training is conducted on 4 GPUs, with a total batch size of 16 across 90k iterations (including 0.5k for burn-in), using a starting learning rate of 0.02, reduced by a factor of 10 at 60k and 80k iterations. The image input size adheres to the standard configuration, with the short side ranging from [640, 800] and the long side under 1333. Additionally, we apply common weak-strong augmentations from semi-supervised object detection iterature~\cite{liu2021unbiased,xu2021end} in pseudo-label learning. For the LVIS dataset, we replicate Detic's experimental setup~\cite{zhou2022detecting} and apply our method to the CenterNet2~\cite{zhou2021probabilistic} baseline. The model is trained on 4 GPUs, while maintaining the total batch size unchanged. All experiments are conducted in Detectron2~\cite{wu2019detectron2}, with further details provided in the supplementary materials.

\begin{table*}[t]\renewcommand\tabcolsep{8pt}
\begin{center}
\newcommand{\tabincell}[2]{\begin{tabular}{@{}#1@{}}#2\end{tabular}}  
  \caption{Comparison with state-of-the-art methods on COCO2017 dataset.}
\scalebox{0.8}{
  \begin{tabular}{c|c|ccc}
  \toprule
    \makebox[0.1\textwidth][c]{Methods}    & \makebox[0.5\textwidth][c]{Training Source} &  \makebox[0.05\textwidth][c]{$AP_{50}^{Novel}$}   &   \makebox[0.05\textwidth][c]{\textcolor{gray}{ $AP_{50}^{Base}$}}    &   \makebox[0.05\textwidth][c]{\textcolor{gray}{ $AP_{50}^{All}$}}    \\
    \hline
	RegionCLIP~\cite{zhong2022regionclip}       & \tabincell{c}{box-level labels in $\mathcal{C}_B$, \\internet sourced image-text pairs,\\pretraining with pseudo box-level labels} &    31.4       & \textcolor{gray}{57.1}  & \textcolor{gray}{50.4}\\
    \hline
    Gao \textit{et al.}~\cite{gao2022open}& \tabincell{c}{box-level labels in $\mathcal{C}_B$, \\internet sourced image-text pairs,\\pseudo-box labels in $\mathcal{C}_N$ generated by ALBEF} &    30.8       & \textcolor{gray}{46.1}  & \textcolor{gray}{42.1}\\
     \hline
	PromptDet~\cite{feng2022promptdet}	       & \tabincell{c}{box-level labels in $\mathcal{C}_B$, \\internet sourced image-text pairs,\\pseudo-box labels in $\mathcal{C}_N$ generated by CLIP} &    26.6       & \textcolor{gray}{-}  & \textcolor{gray}{50.6}\\
  \hline
	OADP~\cite{wang2023object}	       & \tabincell{c}{box-level labels in $\mathcal{C}_B$, \\knowledge distillation from CLIP,\\pseudo-box labels in $\mathcal{C}_N$ generated by CLIP} &    35.6       & \textcolor{gray}{55.8}  & \textcolor{gray}{50.5}\\
    \hline
    Rasheed \textit{et al.}~\cite{bangalath2022bridging}& \tabincell{c}{box-level labels in $\mathcal{C}_B$, \\internet sourced image-text pairs,\\image-level labels for $\mathcal{C}_B\cup\mathcal{C}_N$, pseudo-box labels in $\mathcal{C}_N$} &    36.6       & \textcolor{gray}{54.0}  & \textcolor{gray}{49.4}\\
    
    \hline
\tabincell{c}{VL-PLM~\cite{zhao2022exploiting}\\MarvelOVD(Ours)} & \tabincell{c}{box-level labels in $\mathcal{C}_B$,\\box-level pseudo-labels in $\mathcal{C}_N$ generated with CLIP } & \tabincell{c}{32.3\\ \textbf{38.9}} & \tabincell{c}{\textcolor{gray}{54.0} \\ \textcolor{gray}{56.4}} &  \tabincell{c}{\textcolor{gray}{48.3} \\ \textcolor{gray}{51.8}}\\
  \bottomrule
  \end{tabular}
}
  \label{tab1}
\end{center}
\end{table*}

\subsection{Main Results}

Our approach integrates the detector to refine pseudo-label generation and the subsequent learning phase, significantly reducing noises in pseudo-labels and training boxes. As shown in Table~\ref{tab1}, our method outperforms the baseline method~\cite{zhao2022exploiting} substantially in inferring both base and novel categories. The improvement in base categories stems from stratified label assignment, ensuring undisturbed learning of base categories despite pseudo-labels. The detector's integration effectively mitigates CLIP's inability in distinguishing noise in localized regions, enhancing pseudo-label quality and, in turn, the detector's ability to identify novel objects. Our method is also compared with other state-of-the-art open-vocabulary detection techniques utilizing pseudo-labeling in Table~\ref{tab1}. While existing methods often rely on auxiliary data or supervision, like internet-sourced image-text pairs~\cite{gao2022open,feng2022promptdet}, pseudo-region-text pair pre-training~\cite{zhong2022regionclip}, or auxiliary image-level labels~\cite{bangalath2022bridging}, our approach addresses and resolves fundamental issues in pseudo-label generation and conventional training designs~\cite{he2017mask}, achieving significant gains without extra data or supervision. Results on the LVIS dataset (Table~\ref{tab2}) 
compares our method against the common CenterNet2 baseline~\cite{zhou2022detecting}. Contrary to existing methods~\cite{zhou2022detecting,bangalath2022bridging} that use additional classification data with image-level labels for enhanced novel object detection, our method exploits potential novel objects from original training data. The results in Table~\ref{tab2} indicate that our method also excels in large-scale label spaces.

\begin{table}[t]\renewcommand\tabcolsep{8pt}
\begin{center}
\newcommand{\tabincell}[2]{\begin{tabular}{@{}#1@{}}#2\end{tabular}}  
  \caption{Comparison with state-of-the-art methods on LVIS-v1 dataset. All the methods are derived under the same base detector and experimental settings.} 
  \begin{tabular}{c|cccc}
  \toprule
    Methods    &   $AP_{r}$   &   \textcolor{gray}{ $AP_{c}$}    &    \textcolor{gray}{ $AP_{f}$}  & \textcolor{gray}{AP}    \\
    \midrule
	VLDet~\cite{lin2022learning}       &     22.4       & \textcolor{gray}{-}  & \textcolor{gray}{-}&\textcolor{gray}{34.4}\\
     Detic~\cite{zhou2022detecting}       &     24.6       & \textcolor{gray}{32.5}  & \textcolor{gray}{35.6}&\textcolor{gray}{32.4}\\
	Rasheed \textit{et al.}~\cite{bangalath2022bridging}       &     25.2       & \textcolor{gray}{33.4}  & \textcolor{gray}{35.8}&\textcolor{gray}{32.9}\\
    \rowcolor{blue!5}MarvelOVD(Ours)  & \textbf{26.0} & \textcolor{gray}{34.2}&\textcolor{gray}{36.9}&\textcolor{gray}{34.2}\\
  \bottomrule
  \end{tabular}
  \label{tab2}
\end{center}
\end{table}

\section{Ablation Study}

We conduct experiments on the COCO dataset to assess the effectiveness of our method's key components, with additional ablations detailed in supplementary materials.

\subsection{Effects of Each Component}
Table~\ref{tab3} demonstrates the contribution of each proposed algorithmic component to the final performance. 
Initially, we change the global training setting of VL-PLM~\cite{zhao2022exploiting}, including overall novel loss weight $\gamma$ and the data augmentations. A larger novel loss weight doesn't affect the performance while Weak-Strong augmentations enhance the learning of pseudo-labels. We then implement stratified label assignment to resolve conflicts between novel pseudo-labels and base annotations. This adjustment restores base category detection to supervised performance levels without impacting novel category detection. After that, we introduce online object mining to purify the pseudo labels, which brings a significant improvement in detection accuracy on the novel categories, indicating the effectiveness of our method in offsetting CLIP's localized limitations by leveraging the detector's capabilities. Based on it, applying adaptive proposal reweighting further enhances the average precision for novel categories. The promotion comes from the independent weights computed by adaptive reweighting, which enforces the model to focus on boxes with larger overlaps with actual novel objects. In summary, our method offers a less biased pipeline for pseudo-label-based novel concept learning, which not only effectively purifies the training targets but also optimizes the learning procedure, and significantly enhances performance in both base and novel categories without requiring extra data or pretraining.

\begin{table*}[t]\renewcommand\tabcolsep{8pt}
\begin{center}
\newcommand{\tabincell}[2]{\begin{tabular}{@{}#1@{}}#2\end{tabular}}  
  \caption{Roadmap from existing method to our framework.}

  \begin{tabular}{l|ccc}
  \toprule
        &   $AP_{50}^{Novel}$   &   $AP_{50}^{Base}$    &    $AP_{50}^{All}$ \\
    \midrule
	Supervised by base annotations & - & 56.4 & -\\
	\tabincell{c}{VL-PLM~\cite{zhao2022exploiting}}     &     32.7       & 54.0  & 48.5\\
	\tabincell{c}{VL-PLM(set $\gamma=2$)~\cite{zhao2022exploiting}}     &     32.5       & 54.0  & 48.4\\
	\tabincell{l}{$+$Weak-Strong augmentation}     & 34.2   & 53.9  & 49.1\\ 
     \tabincell{l}{$+$\textit{Stratified Label Assignment}}      & 34.4  & \textit{56.4}$\uparrow$  & 50.5\\ 
	
     \tabincell{l}{$+$\textit{Online object Mining}}     & \textit{37.8}$\uparrow$   & 56.5  & 51.3\\ 
	\tabincell{l}{$+$\textit{Adaptive Proposal Reweighting}}     & \textit{38.9}$\uparrow$   & 56.6  & 51.8\\ 
  \bottomrule
  \end{tabular}

  \label{tab3}
\end{center}

\end{table*}

\subsection{Analysis of Pseudo Labels}

\begin{figure}[t]
  \centering
  \includegraphics[width=\linewidth]{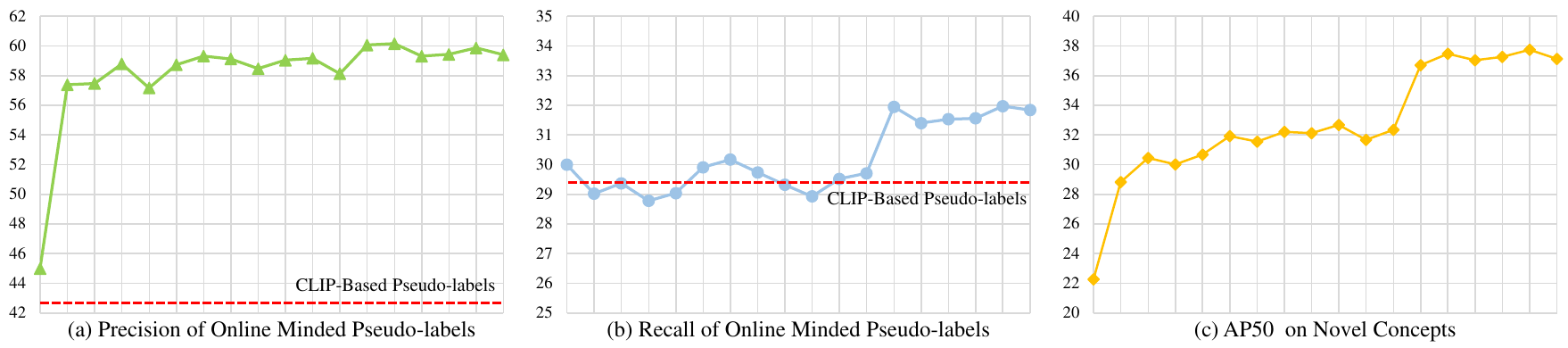}
  \caption{Visualization of the quality of our dynamically generated pseudo-labels, with red dashed lines indicating the quality of the original CLIP-based pseudo-labels.}
  \label{fig_pl}
\end{figure}

We conduct an in-depth analysis of pseudo-labeling quality across different training stages using the COCO evaluation set, which comprises 2064 images featuring novel objects. Pseudo-labels with an IoU score above 0.5 relative to the ground truth (GT) novel object and correctly categorized are classified as True Positives (TPs). The findings are depicted in Figure~\ref{fig_pl}. For equitable comparison, we align the threshold of CLIP-based pseudo-labels ($\delta=0.95$) with our dynamic pseudo-labels ($\delta=0.8$) to maintain comparable recall rates. Initially, the burn-in stage imparts novel object discrimination ability, leading to a stable improvement in pseudo-label quality right after this phase. Subsequently, the enhanced pseudo-labels refine the detector's ability to recognize novel objects, manifesting as a large increase in pseudo-label precision in the early stages of training. As illustrated in Figure~\ref{fig_pl}, as training progresses, the detector increasingly differentiates novel boxes from background and base objects, dynamically enhancing pseudo-label quality. 

\subsection{Effects of different thresholds and burn-in steps}

\begin{table}[t]\renewcommand\tabcolsep{4pt}
\begin{center}
\newcommand{\tabincell}[2]{\begin{tabular}{@{}#1@{}}#2\end{tabular}}  
  \caption{Effects of different thresholds and burn-in steps. The default setting is $\omega=0.5k$ for burn-in and threshold $\delta=0.9$ for online pseudo-label mining.}
  \begin{tabular}{c|cccc||c|cccc}
  \toprule
	$\delta$ & 0.8   & 0.85  & 0.9 & 0.95 & $\omega$ & 0.5k & 1k & 2k & 5k\\
\midrule
	$AP_{50}^{Novel}$ & 37.0 &  38.2 & 38.9 & 38.4 & $AP_{50}^{Novel}$ & 38.9 & 38.7 &  38.7 & 38.5\\
  \bottomrule
  \end{tabular}
  \label{tab5}
\end{center}
\end{table}

We evaluate the performance impact of various thresholds and burn-in steps, detailed in Table~\ref{tab5}. Our base setting uses a threshold $\delta=0.9$ and $\omega=0.5k$ burn-in steps. Notably, the threshold for pseudo-label selection markedly affects model performance; while 0.8 is optimal for our baseline, a threshold of about 0.9 proves more effective for our method due to less biased pseudo-labeling confidence. Regarding burn-in steps, which guide initial learning from CLIP-generated pseudo-labels, their effect on final performance is minimal. As the model converges, the quality of these pseudo-labels improves, indicating that different initial settings eventually yield similar performance outcomes.

\subsection{Dependency analysis of $\lambda$ and $\lambda^\prime$}

\begin{figure}[t]
  \centering
    \includegraphics[width=\linewidth]{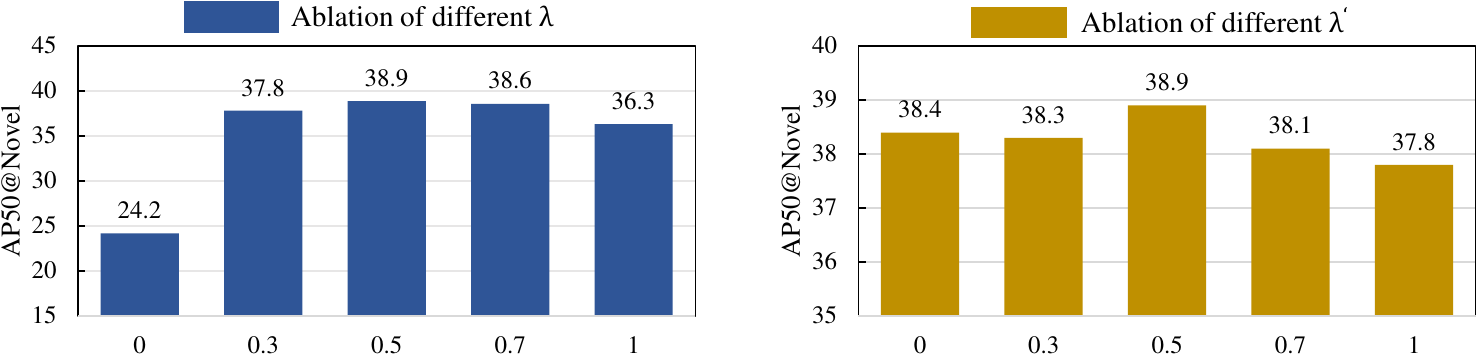}
  \caption{Effects of different dependence controler $\lambda$ and $\lambda^\prime$.}
  \label{fig_lambda}
\end{figure}

We calculate the confidence of each candidate pseudo-label using Eq.~\ref{eq4} and the training weights for each novel box via Eq.~\ref{eq6}, where $\lambda$ and $\lambda^\prime$ determines the reliance on different models or measurements. 
The effects of varying $\lambda$ and $\lambda^\prime$ are documented in Figure~\ref{fig_lambda}. Optimal performance is observed both at $0.5$, and a range of $[0.3,0.7]$ yields comparable outcomes. 
Specifically, extreme values were also tested: $\lambda=0$ implies reliance solely on the detector post burn-in, which results in poor performance, underscoring the importance of CLIP's role in distinguishing novel categories. Conversely, at $\lambda=1$, pseudo-label generation reverts to conventional methods~\cite{zhao2022exploiting, gao2022open, feng2022promptdet} that rely entirely on the CLIP model. While performance decreases in this setting, it still surpasses the baseline, suggesting that adaptive proposal reweighting effectively counters the impact of noisy boxes. By setting $\lambda^\prime=1$, adaptive reweighting changes to the original training design~\cite{he2017mask} with weighted pseudo-labels~\cite{zhao2022exploiting}, and it limits the model's learning of novel concepts, resulting in significant performance degradation.

\section{Conclusion}

In this paper, we address the limitations of pre-trained Vision-Language Models (VLMs) in generating accurate pseudo-labels for localized regions by integrating object detectors' capabilities. The key issue with VLMs is their lack of contextual awareness and inability to differentiate ``background'', leading to biased pseudo-labels. By leveraging the detector's contextual feature extraction and background discrimination abilities, we significantly improve pseudo-label quality through online object mining and optimize the learning process with adaptive proposal reweighting. Our extensive experiments show that this approach not only enhances the detector's novel object recognition but also outperforms state-of-the-art methods without additional data or supervision, offering an efficient and effective solution for learning open vocabulary concepts by pseudo-labels.

\paragraph{\textbf{Acknowledgments.}} This work was supported in part by the Key-Area Research and Development Program of Guangdong Province (NO.~2021B0101420004), in part by the National Natural Science Foundation of China (NO.~62322608, NO.~62106235, NO.~62325605), in part by the Guangxi Science and Technology Plan Project (NO.~GuikeAD23026034), in part by the Shenzhen Science and Technology Program (NO.~JCYJ20220530141211024), and in part by the Open Project Program of the Key Laboratory of Artificial Intelligence for Perception and Understanding, Liaoning Province (AIPU, No.~20230003).

\appendix


\section{Further Ablation Studies}
\label{sec:rationale}

\subsection{Different indicators for adaptive reweighting}

\begin{table}[t]\renewcommand\tabcolsep{4pt}
\begin{center}
\newcommand{\tabincell}[2]{\begin{tabular}{@{}#1@{}}#2\end{tabular}}  
  \caption{Performance of different measurements for reliability score in adaptive proposal reweighting.}
  \begin{tabular}{c|cccc}
  \toprule
    $r_i$ & $1-b_i$  &  $s_i$  &  $iou_i$  & $s_i^{det}$ \\
    \midrule
	$AP_{50}^{Novel}$ & 39.8  & 37.8 & 37.6 & 38.0\\
  \bottomrule
  \end{tabular}
  \label{tab6}
\end{center}
\end{table}

We examine several other measurements to estimate the reliability score $r_i$ for adaptive reweighting, including:
\begin{itemize}
\item \textit{Confidence of pseudo label ($r_i=s_i$)}: For comparsion. By setting $r_i=s_i$, the adaptive reweighting degenerates to conventional training design~\cite{he2017mask} with weighted pseudo-labels~\cite{zhao2022exploiting}.
\item \textit{Intersection-over-Union ($r_i=iou_i$)}: Boxes with bigger overlaps from pseudo-label get larger weights and vice versa.
\item \textit{Novelty estimation ($r_i=s_i^{det}$)}: $s_i^{det}$ defined in Eq.3 in main paper estimates the probability of containing real novel objects, we repeatedly use it for both pseudo-label mining and adaptive reweighting in this setting.
\end{itemize}
As demonstrated in Table~\ref{tab6}, background score predicted with weakly augmented images achieves the best performance, and it significantly outperforms the conventional training design~\cite{he2017mask} with weighted pseudo-labels~\cite{zhao2022exploiting}.

\subsection{Ablation on global novel loss weight $\gamma$}
\begin{table}[h]
\renewcommand\tabcolsep{10pt}
\begin{center}
\newcommand{\tabincell}[2]{\begin{tabular}{@{}#1@{}}#2\end{tabular}} 
  \caption{Effects of different global novel loss weight $\gamma$.}
  \begin{tabular}{c|cc|cc}
  \toprule
	\multirow{2}*{\tabincell{c}{Models}} & \multicolumn{2}{c|}{\underline{VL-PLM}} & \multicolumn{2}{c}{\underline{MarvelOVD}}\\
	 & $AP_{50}^{N}$ & $AP_{50}^{B}$ & $AP_{50}^{N}$ & $AP_{50}^{B}$\\
	\midrule
	$\gamma=1$ & 32.7  &   54.0  &   37.8   &   57.0 \\
	$\gamma=2$ & 32.5  &   53.9  &  \textbf{38.9}   &   56.5 \\
	$\gamma=4$ & -  &   -  &   38.6   &   56.0 \\
  \bottomrule
  \end{tabular}
  \label{tab1}
\end{center}
\vspace{-15pt}
\end{table}
Conventional methods~\cite{zhao2022exploiting, gao2022open, feng2022promptdet} typically treat noisy pseudo-labels as ground truth and combine it with the base annotations to train the detector. In contrast, our proposed MarvelOVD strives to reduce the noises in both pseudo-labels and training boxes before utilizing them in training, which achieves more promising results. We examine different global novel loss weights and report the performance in Table~\ref{tab1}. The results show that setting the global novel weight as $\gamma=2$ further improves the performance of our MarvelOVD while slightly degrading the baseline method. The reason is that the massive noise contained in the baseline method prevents further improvements. On the contrary, our approach effectively purifies the pseudo-labels and de-bias the following training designs by online mining and adaptive reweighting, allowing better performance with larger novel loss weights. Higher novel weights are also tested but do not contribute to better results.

\subsection{Qualitative Results}
\begin{figure}[t]
  \centering
  \includegraphics[width=\linewidth]{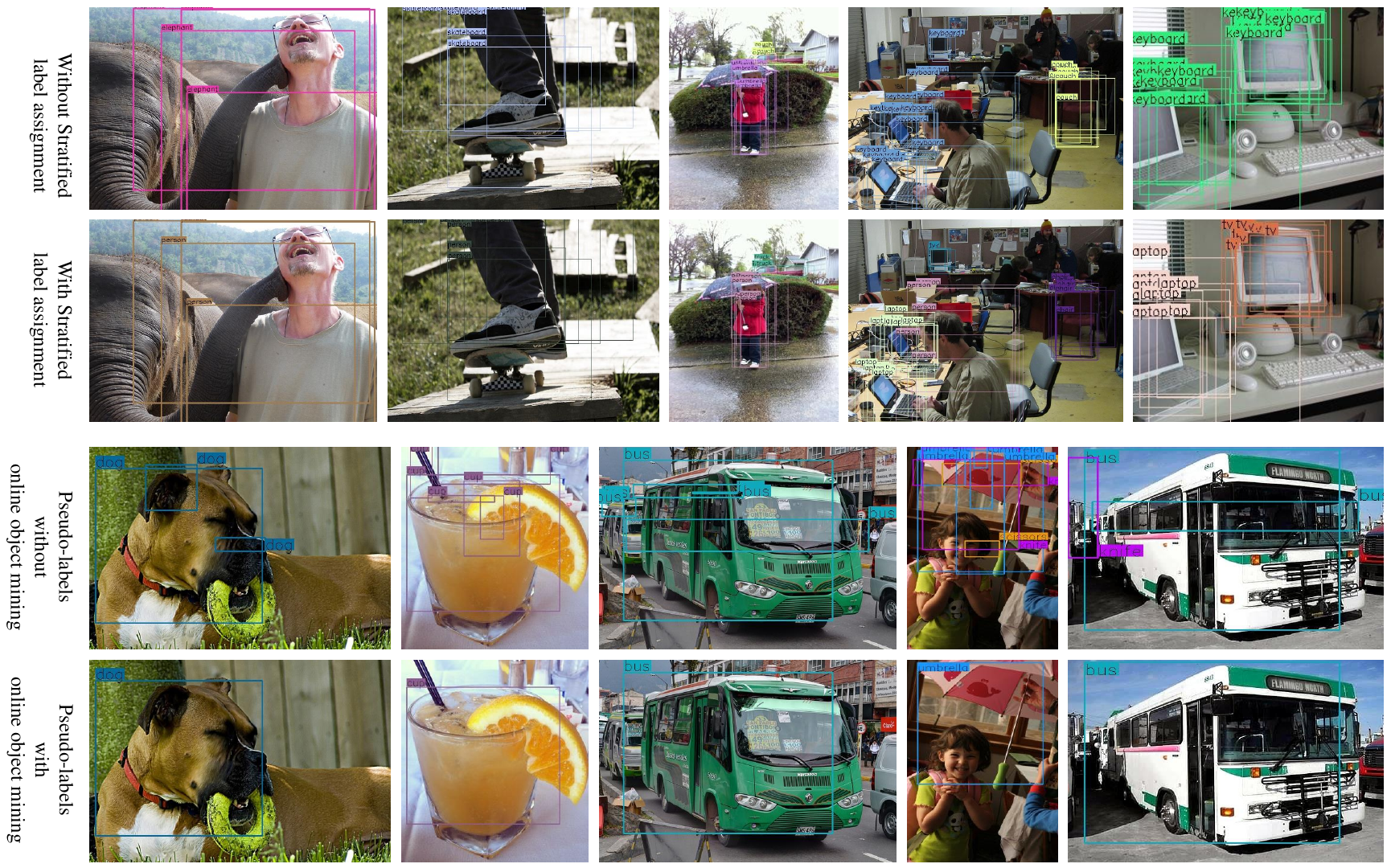}
  \caption{Visualization of stratified label assignment and online object mining.}
  \label{fig_vis}

\end{figure}
In Figure~\ref{fig_vis}, we visualize the effects of our proposed stratified label assignment and online object mining. A notable observation is that base boxes might be incorrectly labeled as novel objects due to overlaps between the pseudo-label and base annotation, potentially impairing the detector's performance on base categories. Our stratified matching strategy rectifies these mislabeled base boxes, enabling the model to assimilate novel concepts without diminishing its base detection capabilities. Additionally, the bottom two rows of Figure~\ref{fig_vis} demonstrate the efficacy of online object mining, where our approach, aided by the detector, effectively filters noise from CLIP-generated pseudo-labels.

\subsection{Effects of the weak-strong augmentation}
\begin{table}[t]\renewcommand\tabcolsep{6pt}
\begin{center}
\newcommand{\tabincell}[2]{\begin{tabular}{@{}#1@{}}#2\end{tabular}}  
  \caption{Effects of Weak-Strong Augmentation~(referred as WSA).}
  \begin{tabular}{c|cc|cc}
  \toprule
	\multirow{2}*{\tabincell{c}{Models}} & \multicolumn{2}{c|}{\underline{VL-PLM}} & \multicolumn{2}{c}{\underline{MarvelOVD}}\\
	 & $AP_{50}^{N}$ & $AP_{50}^{B}$ & $AP_{50}^{N}$ & $AP_{50}^{B}$\\
	\midrule
	w/o WSA & 32.7  &   54.0  &   37.2   &   56.4 \\
	\midrule
	w/ WSA & 34.2  &   53.9  &   38.9   &   56.5 \\
  \bottomrule
  \end{tabular}
  \label{tab2}
\end{center}
\vspace{-10pt}
\end{table}

We apply the weak-strong augmentations that are not adopted by other OVD works. The motivation comes from recent semi-supervised learning methods~(e.g. fixmatch~\cite{sohn2020fixmatch}, unbiased-teacher~\cite{liu2021unbiased}), which demonstrate that enforcing the same supervision between weak-strong augmented features leads to better performance for learning on pseudo-labels. We apply weak-strong augmentations on the conventional OVD method~\cite{zhao2022exploiting} and MarvelOVD to evaluate its effect and the results are shown in Table~\ref{tab2}. The weak-strong augmentation barely influences the average precision on base categories while equally improving the performance on novel categories in both VL-PLM and our MarvelOVD. Without the augmentation, our framework still outperforms the base method by a significant margin.

\begin{table*}[t]\renewcommand\tabcolsep{10pt}
\begin{center}
\newcommand{\tabincell}[2]{\begin{tabular}{@{}#1@{}}#2\end{tabular}}  
  \caption{Comparison with state-of-the-art methods on COCO2017 dataset.}
\scalebox{0.8}{
  \begin{tabular}{c|c|ccc}
  \toprule
    \makebox[0.1\textwidth][c]{Methods}    & \makebox[0.5\textwidth][c]{Training Source} &  \makebox[0.05\textwidth][c]{$AP_{50}^{Novel}$}   &   \makebox[0.05\textwidth][c]{\textcolor{gray}{ $AP_{50}^{Base}$}}    &   \makebox[0.05\textwidth][c]{\textcolor{gray}{ $AP_{50}^{All}$}}    \\ 

\midrule
	
\tabincell{c}{OV-RCNN~\cite{zareian2021open}\\VLDet~\cite{lin2022learning}\\LocOv~\cite{bravo2022localized}} & \tabincell{c}{box-level labels in $\mathcal{C}_B$, \\transfer learning with COCO-captions} & \tabincell{c}{22.8\\32.0\\28.6} & \tabincell{c}{\textcolor{gray}{46.0}\\ \textcolor{gray}{50.6}\\ \textcolor{gray}{51.3}}  & \tabincell{c}{\textcolor{gray}{39.9}\\\textcolor{gray}{45.8}\\\textcolor{gray}{45.7}}\\ 

\midrule
	
Detic~\cite{zhou2022detecting} & \tabincell{c}{box-level labels in $\mathcal{C}_B$, \\internet sourced classification data,\\image-level labels for $\mathcal{C}_B\cup\mathcal{C}_N$} & 27.8 & \textcolor{gray}{47.1} & \textcolor{gray}{45.0}\\ 

\midrule
	
\tabincell{c}{ViLD~\cite{gu2021open}\\BARON~\cite{wu2023aligning}} & \tabincell{c}{box-level labels in $\mathcal{C}_B$, \\knowledge distillation from CLIP} & \tabincell{c}{27.6\\34.0} & \tabincell{c}{\textcolor{gray}{59.5}\\\textcolor{gray}{60.4}} & \tabincell{c}{\textcolor{gray}{51.3}\\\textcolor{gray}{53.5}}\\ 

\midrule

RegionCLIP~\cite{zhong2022regionclip}       & \tabincell{c}{box-level labels in $\mathcal{C}_B$, \\internet sourced image-text pairs,\\pretraining with pseudo box-level labels} &    31.4       & \textcolor{gray}{57.1}  & \textcolor{gray}{50.4}\\

\midrule

Gao \textit{et al.}~\cite{gao2022open}& \tabincell{c}{box-level labels in $\mathcal{C}_B$, \\internet sourced image-text pairs,\\pseudo-box labels in $\mathcal{C}_N$ generated by ALBEF} &    30.8       & \textcolor{gray}{46.1}  & \textcolor{gray}{42.1}\\

\midrule

PromptDet~\cite{feng2022promptdet}	       & \tabincell{c}{box-level labels in $\mathcal{C}_B$, \\internet sourced image-text pairs,\\pseudo-box labels in $\mathcal{C}_N$ generated by CLIP} &    26.6       & \textcolor{gray}{-}  & \textcolor{gray}{50.6}\\

\midrule

OADP~\cite{wang2023object}	       & \tabincell{c}{box-level labels in $\mathcal{C}_B$, \\knowledge distillation from CLIP,\\pseudo-box labels in $\mathcal{C}_N$ generated by CLIP} &    35.6       & \textcolor{gray}{55.8}  & \textcolor{gray}{50.5}\\

\midrule

Rasheed \textit{et al.}~\cite{bangalath2022bridging}& \tabincell{c}{box-level labels in $\mathcal{C}_B$, pseudo-box labels in $\mathcal{C}_N$\\internet sourced image-text pairs,\\image-level labels for $\mathcal{C}_B\cup\mathcal{C}_N$} &    36.6       & \textcolor{gray}{54.0}  & \textcolor{gray}{49.4}\\
    
\midrule

SAS-Det ~\cite{zhao2023taming}& \tabincell{c}{box-level labels in $\mathcal{C}_B$, pseudo-box labels in $\mathcal{C}_N$\\ generated by roi-align from CNN-based-CLIP} &   \underline{37.4}       & \textcolor{gray}{58.0}  & \textcolor{gray}{53.0}\\

\midrule

\tabincell{c}{VL-PLM~\cite{zhao2022exploiting}\\MarvelOVD(Ours)} & \tabincell{c}{box-level labels in $\mathcal{C}_B$,\\box-level pseudo-labels in $\mathcal{C}_N$ generated with CLIP } & \tabincell{c}{32.3\\  \textbf{38.9}} & \tabincell{c}{\textcolor{gray}{54.0} \\ \textcolor{gray}{56.5}} &  \tabincell{c}{\textcolor{gray}{48.3}  \\ \textcolor{gray}{51.9}}\\

  \bottomrule
  \end{tabular}
}
  \label{tab3}
\end{center}
\vspace{-10pt}
\end{table*}

\section{Comparisons and compatibility with existing OVD works}
\label{sec:supp_comps}

\subsection{Performance Comparison}
We mainly compare our method with other pseudo-label-based OVD works in the main paper. Table~\ref{tab3} demonstrates a more complete comparison between our MarvelOVD and other existing OVD works, including both transfer learning and knowledge distillation methods. Among them, our method still performs favorably against the state-of-the-art methods. Pseudo-label plays an important role in recent OVD works, where most advanced methods learn novel concepts from pseudo-labels generated by pretrained VLMs. With respect to VLM-generated pseudo labels, our MarvelOVD identifies the root causes of its noises and proposes the dedicated noise-removal strategy by integrating the context-sensing capability of the detector, which consistently improves the recent advanced method by significant margins. 

In particular, extracting CLIP embedding by roi-align from CNN-based CLIP backbones has been exploited in recent studies~\cite{kuo2022f,yu2024convolutions}, which provides a substituting context-aware operation for pseudo-label generation. Even though, our approach still stably outperforms methods that exploit such operation~(e.g. SAS-Det~\cite{zhao2023taming} that recently published on arxiv). The result further indicates the effectiveness of our MarvelOVD in de-noising the pseudo-label-based learning paradigms.

\section{More Implementation Details}
\label{sec:supp_imp}

\begin{table*}[t]\renewcommand\tabcolsep{10pt}
\begin{center}
\newcommand{\tabincell}[2]{\begin{tabular}{@{}#1@{}}#2\end{tabular}}  
  \caption{Detail of data augmentations. Probability in the table indicates the probability of applying the corresponding image process.}
\scalebox{0.8}{
  \begin{tabular}{ccc}
  \toprule
	\midrule
	\multicolumn{3}{c}{\textbf{Weak Augmentation}}\\ \midrule
	Process & Probability & Parameters\\ \midrule
	Horizontal Flip & 0.5 & None \\ \midrule
	\midrule
	\multicolumn{3}{c}{\textbf{Strong Augmentation}}\\ \midrule
	Process & Probability & Parameters\\ \midrule
	Color Jittering & 0.8 & \tabincell{c}{(brightness, contrast, saturation, hue) = (0.4, 0.4, 0.4, 0.1)}\\ \midrule
	Grayscale & 0.2 & None\\ \midrule
	GaussianBlur & 0.5 &  (sigma\_x, sigma\_y) = (0.1, 2.0)\\ \midrule
	CutoutPattern1 &  0.7 & scale=(0.05, 0.2), ratio=(0.3, 3.3)\\ \midrule
	CutoutPattern2 &  0.5 & scale=(0.02, 0.2), ratio=(0.1, 6)\\ \midrule
	CutoutPattern3 &  0.3 & scale=(0.02, 0.2), ratio=(0.05, 8)\\ \midrule
  \bottomrule
  \end{tabular}
}
  \label{tab5}
\end{center}
\vspace{-10pt}
\end{table*}

\subsection{Weak-Strong augmentations}
The detailed weak-strong augmentations adopted in our method are illustrated in Table~\ref{tab5}, which is identical with the semi-supervised object detection work unbiased teacher~\cite{liu2021unbiased}. We only apply it to the COCO dataset. The augmentation on the LVIS dataset follows the common CenterNet2 benchmark~\cite{zhou2022detecting}.

\subsection{Candidate pseudo-label assignment}

We follow the pseudo-label-generation pipeline in VL-PLM to assign candidate pseudo-labels to each image before the training. In particular, the class-agnostic proposal generator is actually a detector trained with the base annotation~(regarding all the base annotations as one class). Then we infer the train image with the proposal generator and recursively refine the predicted boxes with the RoI head by 10 times. The recursive refinement improves the localization quality of the boxes. After gathering the candidate regions, the prediction probability distribution $\boldsymbol{p}_i$ for each box is encoded by CLIP ViT-B/32 as follows:
\begin{equation}
\label{eq:loss2}
\begin{aligned}
\boldsymbol{r}_i &= \phi(E_{img}(R_i^{1\times}) + E_{img}(R_i^{1.5\times}))\\
\boldsymbol{p}_i = so&ftmax\{\boldsymbol{r}_i \cdot E_{txt}(Novel Categories)^T\}
\end{aligned}
\end{equation}
$E_{img},E_{txt}$ is the image-encoder and text-encoder of CLIP, $R_i^{1\times}$ is the box produced by proposal generator and $R_i^{1.5\times}$ is a region cropped by $1.5\times$ the size of $R_i^{1\times}$. 
After getting the probability distribution for each box, we filter them with a threshold 0.5 and post-process the remaining with NMS to obtain the candidate pseudo-labels. We record the TOP-1 CLIP score and the predicted category of the candidates and assign them to the image, and then we dynamically select reliable ones for training under the guidance of detector. The threshold 0.5 is not a special hyperparameter that influences the performance, it's used to remove the redundant boxes that would never be selected as pseudo-labels, which accelerates the training speeds. Refining the localization with the RoI head and extracting the region-embedding with a larger area are existing techniques that adopted by the base method VL-PLM. We also maintain them in our candidate pseudo-label assignment process.

\section{Limitations}
\label{sec:limits}

Our proposed MarvelOVD provides a better measurement to purify pseudo-labels from the fixed candidate boxes. It can not promote the localization quality of the pseudo-label. Since the candidate boxes are produced by a proposal generator trained with only base annotations, the localization quality for novel objects is limited. As the detector gains more knowledge of the novel object through pseudo-labels during training, its ability to localize the novel object should also be enhanced. How to rationally utilize the detector to dynamically optimize the localization quality of pseudo-labels is worthwhile exploring in the future.

%
%
\bibliographystyle{splncs04}
\bibliography{main}

\begin{thebibliography}{10}
\providecommand{\url}[1]{\texttt{#1}}
\providecommand{\urlprefix}{URL }
\providecommand{\doi}[1]{https://doi.org/#1}

\bibitem{bangalath2022bridging}
Bangalath, H., Maaz, M., Khattak, M.U., Khan, S.H., Shahbaz~Khan, F.: Bridging
  the gap between object and image-level representations for open-vocabulary
  detection. Advances in Neural Information Processing Systems  \textbf{35},
  33781--33794 (2022)

\bibitem{bravo2022localized}
Bravo, M.A., Mittal, S., Brox, T.: Localized vision-language matching for
  open-vocabulary object detection. In: DAGM German Conference on Pattern
  Recognition. pp. 393--408. Springer (2022)

\bibitem{chen2022open}
Chen, P., Sheng, K., Zhang, M., Lin, M., Shen, Y., Lin, S., Ren, B., Li, K.:
  Open vocabulary object detection with proposal mining and prediction
  equalization. arXiv preprint arXiv:2206.11134  (2022)

\bibitem{du2022learning}
Du, Y., Wei, F., Zhang, Z., Shi, M., Gao, Y., Li, G.: Learning to prompt for
  open-vocabulary object detection with vision-language model. In: Proceedings
  of the IEEE/CVF Conference on Computer Vision and Pattern Recognition. pp.
  14084--14093 (2022)

\bibitem{feng2022promptdet}
Feng, C., Zhong, Y., Jie, Z., Chu, X., Ren, H., Wei, X., Xie, W., Ma, L.:
  Promptdet: Towards open-vocabulary detection using uncurated images. In:
  European Conference on Computer Vision. pp. 701--717. Springer (2022)

\bibitem{frome2013devise}
Frome, A., Corrado, G.S., Shlens, J., Bengio, S., Dean, J., Ranzato, M.,
  Mikolov, T.: Devise: A deep visual-semantic embedding model. Advances in
  neural information processing systems  \textbf{26} (2013)

\bibitem{gao2022open}
Gao, M., Xing, C., Niebles, J.C., Li, J., Xu, R., Liu, W., Xiong, C.: Open
  vocabulary object detection with pseudo bounding-box labels. In: European
  Conference on Computer Vision. pp. 266--282. Springer (2022)

\bibitem{gu2021open}
Gu, X., Lin, T.Y., Kuo, W., Cui, Y.: Open-vocabulary object detection via
  vision and language knowledge distillation. arXiv preprint arXiv:2104.13921
  (2021)

\bibitem{gupta2019lvis}
Gupta, A., Dollar, P., Girshick, R.: Lvis: A dataset for large vocabulary
  instance segmentation. In: Proceedings of the IEEE/CVF conference on computer
  vision and pattern recognition. pp. 5356--5364 (2019)

\bibitem{he2017mask}
He, K., Gkioxari, G., Doll{\'a}r, P., Girshick, R.: Mask r-cnn. In: Proceedings
  of the IEEE international conference on computer vision. pp. 2961--2969
  (2017)

\bibitem{he2016deep}
He, K., Zhang, X., Ren, S., Sun, J.: Deep residual learning for image
  recognition. In: Proceedings of the IEEE conference on computer vision and
  pattern recognition. pp. 770--778 (2016)

\bibitem{jayaraman2014zero}
Jayaraman, D., Grauman, K.: Zero-shot recognition with unreliable attributes.
  Advances in neural information processing systems  \textbf{27} (2014)

\bibitem{jia2021scaling}
Jia, C., Yang, Y., Xia, Y., Chen, Y.T., Parekh, Z., Pham, H., Le, Q., Sung,
  Y.H., Li, Z., Duerig, T.: Scaling up visual and vision-language
  representation learning with noisy text supervision. In: International
  conference on machine learning. pp. 4904--4916. PMLR (2021)

\bibitem{kim2021vilt}
Kim, W., Son, B., Kim, I.: Vilt: Vision-and-language transformer without
  convolution or region supervision. In: International Conference on Machine
  Learning. pp. 5583--5594. PMLR (2021)

\bibitem{kuo2022f}
Kuo, W., Cui, Y., Gu, X., Piergiovanni, A., Angelova, A.: F-vlm:
  Open-vocabulary object detection upon frozen vision and language models.
  arXiv preprint arXiv:2209.15639  (2022)

\bibitem{li2022blip}
Li, J., Li, D., Xiong, C., Hoi, S.: Blip: Bootstrapping language-image
  pre-training for unified vision-language understanding and generation. In:
  International Conference on Machine Learning. pp. 12888--12900. PMLR (2022)

\bibitem{li2021align}
Li, J., Selvaraju, R., Gotmare, A., Joty, S., Xiong, C., Hoi, S.C.H.: Align
  before fuse: Vision and language representation learning with momentum
  distillation. Advances in neural information processing systems  \textbf{34},
   9694--9705 (2021)

\bibitem{li2023distilling}
Li, L., Miao, J., Shi, D., Tan, W., Ren, Y., Yang, Y., Pu, S.: Distilling detr
  with visual-linguistic knowledge for open-vocabulary object detection. In:
  Proceedings of the IEEE/CVF International Conference on Computer Vision. pp.
  6501--6510 (2023)

\bibitem{li2022grounded}
Li, L.H., Zhang, P., Zhang, H., Yang, J., Li, C., Zhong, Y., Wang, L., Yuan,
  L., Zhang, L., Hwang, J.N., et~al.: Grounded language-image pre-training. In:
  Proceedings of the IEEE/CVF Conference on Computer Vision and Pattern
  Recognition. pp. 10965--10975 (2022)

\bibitem{lin2022learning}
Lin, C., Sun, P., Jiang, Y., Luo, P., Qu, L., Haffari, G., Yuan, Z., Cai, J.:
  Learning object-language alignments for open-vocabulary object detection.
  arXiv preprint arXiv:2211.14843  (2022)

\bibitem{lin2017feature}
Lin, T.Y., Doll{\'a}r, P., Girshick, R., He, K., Hariharan, B., Belongie, S.:
  Feature pyramid networks for object detection. In: Proceedings of the IEEE
  conference on computer vision and pattern recognition. pp. 2117--2125 (2017)

\bibitem{lin2014microsoft}
Lin, T.Y., Maire, M., Belongie, S., Hays, J., Perona, P., Ramanan, D.,
  Doll{\'a}r, P., Zitnick, C.L.: Microsoft coco: Common objects in context. In:
  Computer Vision--ECCV 2014: 13th European Conference, Zurich, Switzerland,
  September 6-12, 2014, Proceedings, Part V 13. pp. 740--755. Springer (2014)

\bibitem{liu2021unbiased}
Liu, Y.C., Ma, C.Y., He, Z., Kuo, C.W., Chen, K., Zhang, P., Wu, B., Kira, Z.,
  Vajda, P.: Unbiased teacher for semi-supervised object detection. In:
  Proceedings of the International Conference on Learning Representations
  (ICLR) (2021)

\bibitem{lu2019vilbert}
Lu, J., Batra, D., Parikh, D., Lee, S.: Vilbert: Pretraining task-agnostic
  visiolinguistic representations for vision-and-language tasks. Advances in
  neural information processing systems  \textbf{32} (2019)

\bibitem{radford2021learning}
Radford, A., Kim, J.W., Hallacy, C., Ramesh, A., Goh, G., Agarwal, S., Sastry,
  G., Askell, A., Mishkin, P., Clark, J., et~al.: Learning transferable visual
  models from natural language supervision. In: International conference on
  machine learning. pp. 8748--8763. PMLR (2021)

\bibitem{rao2022denseclip}
Rao, Y., Zhao, W., Chen, G., Tang, Y., Zhu, Z., Huang, G., Zhou, J., Lu, J.:
  Denseclip: Language-guided dense prediction with context-aware prompting. In:
  Proceedings of the IEEE/CVF Conference on Computer Vision and Pattern
  Recognition. pp. 18082--18091 (2022)

\bibitem{shi2022proposalclip}
Shi, H., Hayat, M., Wu, Y., Cai, J.: Proposalclip: Unsupervised open-category
  object proposal generation via exploiting clip cues. In: Proceedings of the
  IEEE/CVF Conference on Computer Vision and Pattern Recognition. pp.
  9611--9620 (2022)

\bibitem{sohn2020fixmatch}
Sohn, K., Berthelot, D., Carlini, N., Zhang, Z., Zhang, H., Raffel, C.A.,
  Cubuk, E.D., Kurakin, A., Li, C.L.: Fixmatch: Simplifying semi-supervised
  learning with consistency and confidence. Advances in neural information
  processing systems  \textbf{33},  596--608 (2020)

\bibitem{wang2023object}
Wang, L., Liu, Y., Du, P., Ding, Z., Liao, Y., Qi, Q., Chen, B., Liu, S.:
  Object-aware distillation pyramid for open-vocabulary object detection. In:
  Proceedings of the IEEE/CVF Conference on Computer Vision and Pattern
  Recognition. pp. 11186--11196 (2023)

\bibitem{wu2023aligning}
Wu, S., Zhang, W., Jin, S., Liu, W., Loy, C.C.: Aligning bag of regions for
  open-vocabulary object detection. In: Proceedings of the IEEE/CVF Conference
  on Computer Vision and Pattern Recognition. pp. 15254--15264 (2023)

\bibitem{wu2023cora}
Wu, X., Zhu, F., Zhao, R., Li, H.: Cora: Adapting clip for open-vocabulary
  detection with region prompting and anchor pre-matching. In: Proceedings of
  the IEEE/CVF Conference on Computer Vision and Pattern Recognition. pp.
  7031--7040 (2023)

\bibitem{wu2019detectron2}
Wu, Y., Kirillov, A., Massa, F., Lo, W.Y., Girshick, R.: Detectron2.
  \url{https://github.com/facebookresearch/detectron2} (2019)

\bibitem{xu2021end}
Xu, M., Zhang, Z., Hu, H., Wang, J., Wang, L., Wei, F., Bai, X., Liu, Z.:
  End-to-end semi-supervised object detection with soft teacher. In:
  Proceedings of the IEEE/CVF International Conference on Computer Vision. pp.
  3060--3069 (2021)

\bibitem{yu2024convolutions}
Yu, Q., He, J., Deng, X., Shen, X., Chen, L.C.: Convolutions die hard:
  Open-vocabulary segmentation with single frozen convolutional clip. Advances
  in Neural Information Processing Systems  \textbf{36} (2024)

\bibitem{yun2023ifseg}
Yun, S., Park, S.H., Seo, P.H., Shin, J.: Ifseg: Image-free semantic
  segmentation via vision-language model. In: Proceedings of the IEEE/CVF
  Conference on Computer Vision and Pattern Recognition. pp. 2967--2977 (2023)

\bibitem{zang2022open}
Zang, Y., Li, W., Zhou, K., Huang, C., Loy, C.C.: Open-vocabulary detr with
  conditional matching. In: European Conference on Computer Vision. pp.
  106--122. Springer (2022)

\bibitem{zareian2021open}
Zareian, A., Rosa, K.D., Hu, D.H., Chang, S.F.: Open-vocabulary object
  detection using captions. In: Proceedings of the IEEE/CVF Conference on
  Computer Vision and Pattern Recognition. pp. 14393--14402 (2021)

\bibitem{zhai2022lit}
Zhai, X., Wang, X., Mustafa, B., Steiner, A., Keysers, D., Kolesnikov, A.,
  Beyer, L.: Lit: Zero-shot transfer with locked-image text tuning. In:
  Proceedings of the IEEE/CVF Conference on Computer Vision and Pattern
  Recognition. pp. 18123--18133 (2022)

\bibitem{zhang2022glipv2}
Zhang, H., Zhang, P., Hu, X., Chen, Y.C., Li, L., Dai, X., Wang, L., Yuan, L.,
  Hwang, J.N., Gao, J.: Glipv2: Unifying localization and vision-language
  understanding. Advances in Neural Information Processing Systems
  \textbf{35},  36067--36080 (2022)

\bibitem{zhao2023taming}
Zhao, S., Schulter, S., Zhao, L., Zhang, Z., G, V.K.B., Suh, Y., Chandraker,
  M., Metaxas, D.N.: Taming self-training for open-vocabulary object detection
  (2023)

\bibitem{zhao2022exploiting}
Zhao, S., Zhang, Z., Schulter, S., Zhao, L., Vijay~Kumar, B., Stathopoulos, A.,
  Chandraker, M., Metaxas, D.N.: Exploiting unlabeled data with vision and
  language models for object detection. In: European Conference on Computer
  Vision. pp. 159--175. Springer (2022)

\bibitem{zhong2022regionclip}
Zhong, Y., Yang, J., Zhang, P., Li, C., Codella, N., Li, L.H., Zhou, L., Dai,
  X., Yuan, L., Li, Y., et~al.: Regionclip: Region-based language-image
  pretraining. In: Proceedings of the IEEE/CVF Conference on Computer Vision
  and Pattern Recognition. pp. 16793--16803 (2022)

\bibitem{zhou2022extract}
Zhou, C., Loy, C.C., Dai, B.: Extract free dense labels from clip. In: European
  Conference on Computer Vision. pp. 696--712. Springer (2022)

\bibitem{zhou2022detecting}
Zhou, X., Girdhar, R., Joulin, A., Kr{\"a}henb{\"u}hl, P., Misra, I.: Detecting
  twenty-thousand classes using image-level supervision. In: European
  Conference on Computer Vision. pp. 350--368. Springer (2022)

\bibitem{zhou2021probabilistic}
Zhou, X., Koltun, V., Kr{\"a}henb{\"u}hl, P.: Probabilistic two-stage
  detection. arXiv preprint arXiv:2103.07461  (2021)

\end{thebibliography}
\end{document}